\documentclass[10pt, a4paper]{article}

\usepackage{lrec-coling2024} % this is the new style

\usepackage{times}
\usepackage{latexsym}

\usepackage[T1]{fontenc}
\usepackage[utf8]{inputenc}

\usepackage{microtype}

\usepackage{algorithm}
\usepackage{algorithmic}
\usepackage{graphicx}
\usepackage{booktabs} 
\usepackage{multicol}
\usepackage{multirow}
\usepackage{amsmath}
\usepackage{amsfonts} 
\usepackage{subfigure}
\usepackage{marvosym}

\usepackage{stfloats}

\usepackage{color}

\newtheorem{assumption}{Assumption}
\newtheorem{problem}{Problem}

\newcommand{\rere}[0]{\textsc{ReRe}}

\newcommand{\nop}[1]{}

\title{Improving Recall of Large Language Models: A Model Collaboration Approach for Relational Triple Extraction\\ \vspace*{.5\baselineskip}}

\name{\begin{tabular}{c}
     Guochao Jiang$^\dag$, Ziqin Luo$^\dag$, Yuchen Shi$^\dag$, Dixuan Wang$^\dag$, \\
     Jiaqing Liang$^{\dag\ddag}$, Deqing Yang$^{\dag\ddag\textrm{\Letter}}$
\end{tabular}}

\name{\begin{tabular}{c}
Zepeng Ding$^\diamondsuit$$^\heartsuit$, Wenhao Huang$^\diamondsuit$$^\spadesuit$, Jiaqing Liang$^\diamondsuit$$^\heartsuit$\textsuperscript{\Letter}, \\ Deqing Yang$^\diamondsuit$$^\heartsuit$, Yanghua Xiao$^\diamondsuit$$^\spadesuit$
\end{tabular}}

\address{$^\diamondsuit$Shanghai Key Laboratory of Data Science \\$^\heartsuit$School of Data Science, $^\spadesuit$School of Computer Science, Fudan University \\
         \texttt{\{zpding22, whhuang21\}@m.fudan.edu.cn} \\
         \texttt{\{liangjiaqing, yangdeqing, shawyh\}@fudan.edu.cn}\\}

\abstract{
Relation triple extraction, which outputs a set of triples from long sentences, plays a vital role in knowledge acquisition. Large language models can accurately extract triples from simple sentences through few-shot learning or fine-tuning when given appropriate instructions. However, they often miss out when extracting from complex sentences.
In this paper, we design an evaluation-filtering framework that integrates large language models with small models for relational triple extraction tasks. The framework includes an evaluation model that can extract related entity pairs with high precision. We propose a simple labeling principle and a deep neural network to build the model, embedding the outputs as prompts into the extraction process of the large model. We conduct extensive experiments to demonstrate that the proposed method can assist large language models in obtaining more accurate extraction results, especially from complex sentences containing multiple relational triples. Our evaluation model can also be embedded into traditional extraction models to enhance their extraction precision from complex sentences. 
 \\ \newline }
\begin{document}
\maketitleabstract

\section{Introduction}
Relational triple extraction plays an important role in knowledge acquisition.
This task aims at extracting triples \textit{(subject, predicate, object)} (or \textit{(s, p, o)}) from a given natural language sentence.
% With the development of pre-trained language models (PLMs) and advances in language understanding technology, 
Current large language models (LLMs) have demonstrated the capacity to effectively extract triples from simple sentences via zero-shot or few-shot learning~\cite{wei2023zero, wadhwa-etal-2023-revisiting}.
However, it is still unsatisfactory when the sentences contain multiple relational triples or mention many entities and relations. 
% For LLMs, comprehending such a complex sentence is not a challenge, but
When LLMs executing multiple triple extraction tasks, they often miss out triples, which means the \textbf{low recall} of the results (see Table~\ref{tab:lowrecall}).
% Moreover, such complex sentences are non-ignorable sources of many triples~\cite{hacred}, so enhancing the model's capability to extract triples from complex sentences is both meaningful and necessary.
%Information extraction is also an important task to evaluate the model's capability in language understanding [citations]. 
% In this paper, we concentrates on improving the performance of LLMs on the task of \emph{multiple relational triple extraction}, which aims to extract multiple relational triples from a single complex sentence.
%~\{\textit{$(s_1, p_1, o_1), (s_2, p_2, o_2), ...$}\}.
%where $s_i$ and $o_i$ are all sub-strings representing entities or values in the given sentence.

\begin{figure*}
\begin{center}
\subfigure[]{
\includegraphics[width=0.50\linewidth]{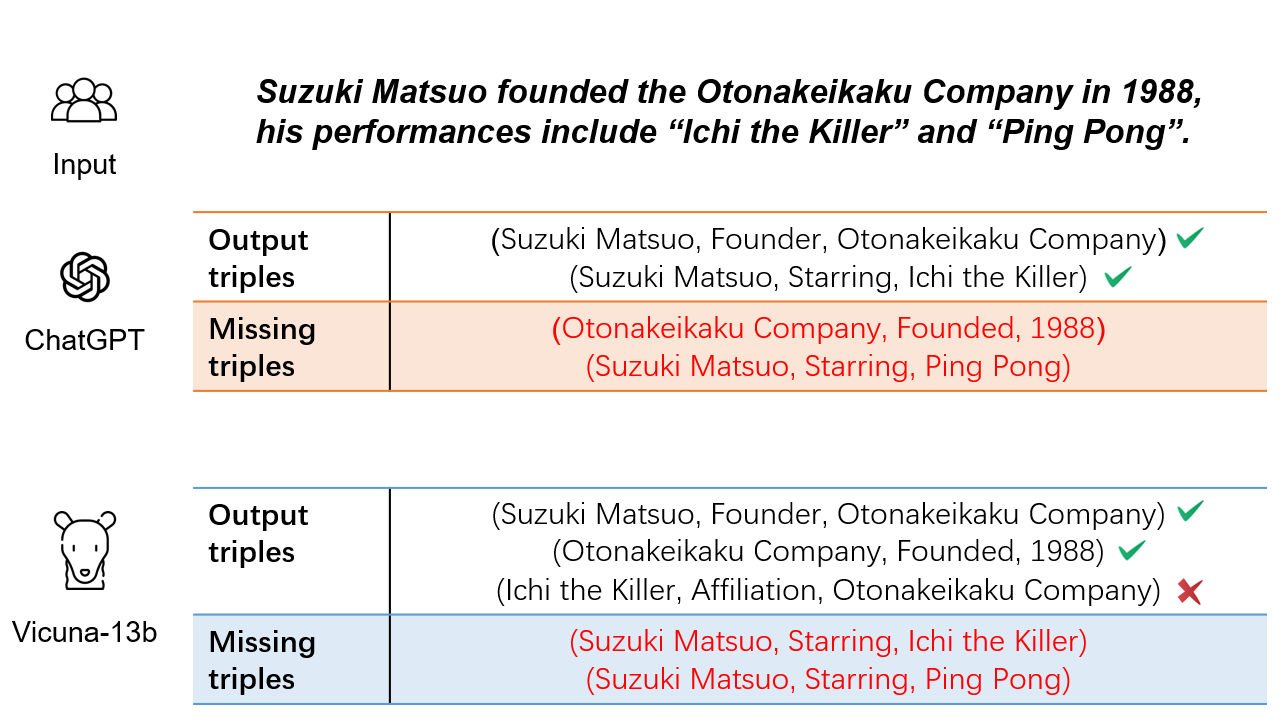}
\label{fig:moti_1}
}
\subfigure[]{
\includegraphics[width=0.4\linewidth]{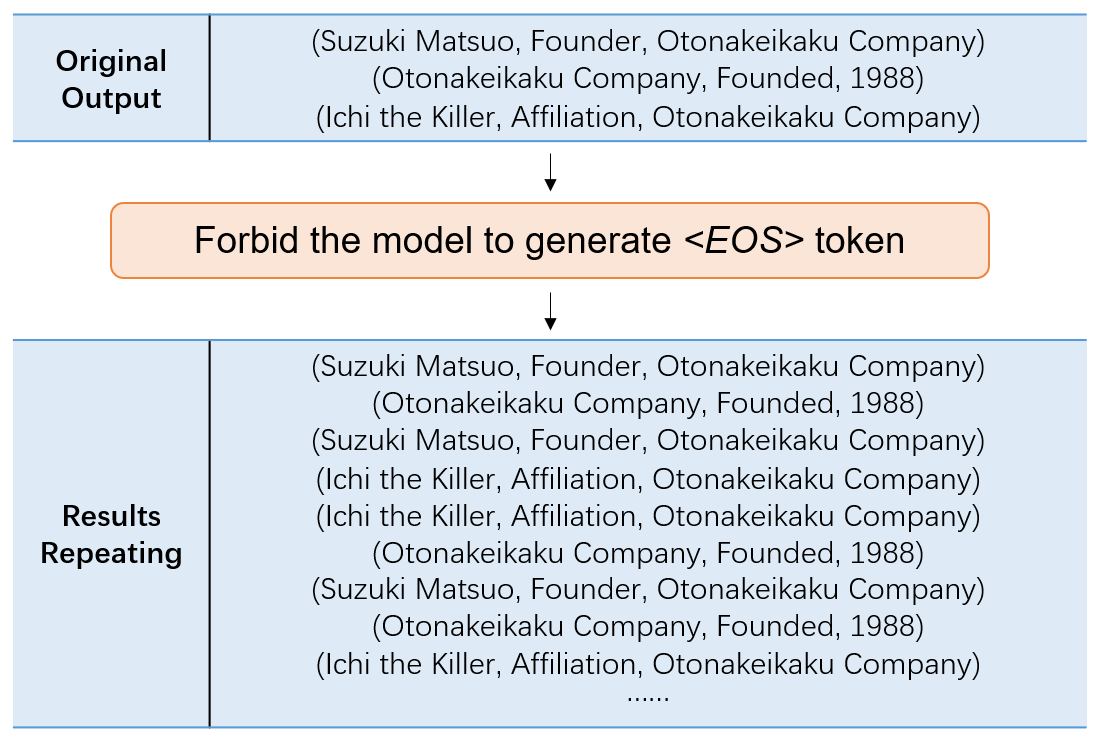}
\label{fig:moti_2}
}
\caption{(a) Illustration of multiple relational triple extraction by LLMs, based on ChatGPT or Vicuna-13B. Both models are given appropriate instructions, limited predicates list and asked to extract as many as possible. (b) Compelling LLM to generate more triples results in repetitive outputs.}
\label{fig:moti}
\end{center}
\end{figure*}

% \begin{figure}[!ht]
% \begin{center}
% \includegraphics[width=1\linewidth]{figures/motivation.png} 
% \caption{Illustration of multiple relational triple extraction by LLMs, based on ChatGPT or Vicuna-13b. Both models are given appropriate instructions, limited predicates list and asked to extract as many as possible.}
% \label{fig:moti}
% \end{center}
% \end{figure}

In the field of relational triple extraction, numerous short sentences contain a substantial number of triples~\cite{hacred}. This presents a significant challenge for LLM-based extraction. Although we can meticulously design instructions or use few-shot in-context learning to improve the triple extraction capabilities of LLMs, it is still difficult to rectify the issue of incomplete extraction from complex sentences by just modifying instructions or incorporating phrases such as `extract as many results as possible' into the prompts, as shown in Figure~\ref{fig:moti_1} and Table~\ref{tab:lowrecall}. This phenomenon could potentially be attributed to the fact that the majority of the training corpus of LLMs is composed of simple sentences, which means the distribution is significantly biased towards containing few triples, leading models to overlook some correct triples when dealing with complex sentences. Previous research has demonstrated that fine-tuning a large model using task data can effectively enhance its relation extraction capabilities, yielding more accurate extraction results~\cite{wadhwa-etal-2023-revisiting}. However, we observed that the fine-tuned LLMs still encounter the issue of a significantly lower recall compared to precision in the case of multiple triples, as Table~\ref{tab:lowrecall} and Table~\ref{tab:mainresult} show. On one hand, the volume of fine-tuning task data is relatively small compared to the original training data of LLMs, making it insufficient to alter the bias of the triplet distribution. On the other hand, the decoding method of the generative language model is not well-suited for the extraction of multiple different relational triples. Even if we compel the large model not to generate the \textit{<EOS>} token unless it produces enough triples, the model still lacks the capability to find more valid triples. Instead, it repeats the generated contents, as shown in Figure~\ref{fig:moti_2}.
% we modify the decoding section of the open-source LLM, such as prohibiting the generation of periods and end symbols, to force it to generate more outputs. 
% As shown in Figure~\ref{fig:moti_2}, the large model does not generate new triples, but instead repeats the generated contents.

\begin{table}[htb]
\small
\centering
	\begin{tabular}{cccc}
		\toprule
		\textbf{Dataset} & \textbf{Fine-tune} & \textbf{Precision} & \textbf{Recall} \\ 
		\midrule 
		NYT10 & w/o fine-tune & 13.75  & 7.75   \\ 
		NYT10 & w/ fine-tune & 78.05  & 46.38  \\ 
		SKE21 & w/o fine-tune & 41.30 &  34.17 \\ 
		SKE21 & w/ fine-tune & 72.56  & 57.42 \\ 
		\bottomrule
	\end{tabular}
	\caption{Precision and recall when extracting multiple relational triples by a large language model. We only consider the complex sentences that contain more than 7 triples. The model used is Vicuna-13B for NYT10 and Qwen-7B for SKE21.}
	\label{tab:lowrecall}
\end{table}

As a result, relying solely on LLMs to achieve complete extraction results in multiple triple extraction tasks is proved to be a considerable challenge.
% while current LLMs excel at extracting simple sentences and achieving high-precision results, their performance in multiple relational triple extraction leaves much to be desired. 
% Notably, the recall of the results is significantly lower than that of regular extraction models. 
% This phenomenon could potentially be attributed to the fact that the majority of training corpus of LLMs is composed of simple sentences, which means the distribution is significantly biased towards containing few triples, leads models to overlook some correct triples when dealing with complex sentence. 
Conversely, traditional small models are prone to extract an excessive number of triples, leading to high recall but low precision results. 
It is because these models lack the ability to identify what triples are not mentioned~\cite{chu2020insrl, jiang2020surface}.

Hence, the model collaboration methods that amalgamate the strengths of small models and LLMs are a natural consideration for addressing the multiple relational triple extractions. 
% However, it is non-trivial to construct such an effective combinatorial extraction model or framework. 
However, as previously mentioned, traditional small models can easily generate incorrect entity pairs when dealing with complex sentences. If these results are directly incorporated into the extraction process, such as providing them to the LLMs as part of the prompt, it can easily mislead the LLMs and compromise extraction precision.
% Therefore, before integrating the extraction results of the small model, an additional filtering processing is necessary to eliminate as many erroneous entity pairs as possible while retaining the positive pairs. Subsequently, these results can be combined with the LLMs to enhance the extraction effect, improving the recall without sacrificing precision.

Motivated by the above considerations, we propose an evaluation-filtering model based on the transformer architecture to generate candidate relational entity pairs and construct an LLMs-based relational triple extraction framework in conjunction with this model.
This model has the following characteristics:
First, the model works at the token level, enabling the evaluation of candidate entity pairs represented with arbitrary tokens, so that it can accurately extract the positive entity pairs and tolerate noisy candidates.
Second, our model can be easily integrated into the process of LLM-based relation triple extraction as a plug-in, significantly enhancing the extraction recall rate. The model can also be seamlessly combined with traditional extraction models to improve the precision.
% Moreover, this evaluation model has a well-designed 2-dim decoder that outputs a matrix, thus it has the ability to evaluate $O(n^2)$ pairs in one inference, to alleviate the inefficiency issue. We propose a self-labeling method, which automatically generates high-quality training samples from existing relational triple extraction datasets, to feed the evaluation models with sufficient high-quality samples.
In summary, our main contributions are as follows:

\begin{itemize}
	\item First, we construct an extraction framework that integrates both the small models and LLMs. This framework provides the filtered positive entity pairs to the LLMs as part of the prompt, thereby guiding the model to consider more entity pairs and assign proper relations to them.
	\item Second, we propose a fast and robust evaluation model that can be used to effectively filter wrong extracted results and generate positive entity pairs.
     It is loosely coupled with the extraction process and can be injected into any small model and LLM-based method to enhance the recall and F1 score of results.
	\item Third, we conduct extensive experiments to show that the proposed method can successfully enhance the performance of LLMs in relational triple extraction, particularly in terms of the recall rate. Additionally, supplementary experiments also indicate that the evaluation-filtering method can boost extraction precision when applied to traditional small models.
\end{itemize}

\section{Related Works}

\subsection{Large Language Models for Relational Triple Extraction}

Large Language Models (LLMs) have gained widespread attention due to their strong ability for various NLP tasks. In addition to the robust GPT series~\cite{brown2020language, openai2023gpt4}, open-source LLMs have been also widely studied and applied, including Llama series~\cite{touvron2023llama, touvron2023llama2}, Qwen~\cite{bai2023qwen} and Vicuna~\cite{zheng2023judging}. Recent studies on LLMs suggest that they perform well in a variety of downstream tasks, even when provided with only a few examples as instructions~\cite{agrawal2022large, jeblick2023chatgpt}. 
In extraction-related tasks, some works show that with proper prompting, ChatGPT can achieve comparable performance with the supervised methods on zero-shot or few-shot settings of extraction tasks~\cite{wei2023zero, gao2023exploring, tang2023does}. For open-source LLMs, previous work shows Flan-T5~\cite{chung2022scaling} can yield outstanding performance by supervising and fine-tuning and suggests LLMs should be a standard baseline for relation extractions~\cite{wadhwa-etal-2023-revisiting}. However, these studies did not specifically consider the model's extraction ability on complex sentences containing multiple relational triples. Furthermore, the manual evaluation of the results was not as rigorous as exact matching, and most of these studies focus on chatGPT and do not consider various open-source LLMs.
% Relational triple extraction is widely studied, but extracting complex sentences is still a challenge.
% Many traditional models cannot handle the complicated triple overlap.
% For example, a generic sequence tagging based method can hardly handle the case where an entity belongs to multiple triples~\cite{Zheng2017JointEO,dai2019joint,tan2019jointly} since most sequence tagging models assume that a token only has one tag.
% A similar difficulty is encountered when multiple triples share the same relation. 
% Most existing tagging mechanism provides insufficient information for the succeeding step to pair relevant entities into proper triples. 
% To overcome the weakness of existing tagging mechanisms, recent approaches~\cite{Cohen2020RelationEA,wang2020tplinker} significantly extend the tagging spaces to model complex overlapping situations, which inevitably suffer from the imbalance of tag distribution and the accompanying difficulties of model training.

\subsection{Model Collaboration in the Era of Large Language Models}

% While most existing end-to-end approaches cannot effectively handle the complex triple set scenarios with a simple decoder, many methods tend to decompose the task into multiple steps and derive the final results accordingly.

Current methods of model collaboration involving large language models can be primarily categorized into three types. First, the output results of the small model are utilized as a component of the overall framework to assist the LLMs to perform better on downstream tasks~\cite{xu2023small, leviathan2023fast}. Second, large and small models are collaboratively trained based on task data to efficiently utilize the unlabeled data and minimize the bias of models~\cite{lang2022co}. Third, ensembling multiple prompts or multiple LLMs to achieve more stable output results, as well as improved generalization performance~\cite{pmlr-v202-allingham23a, jiang-etal-2023-llm}.

In the field of relational triple extraction, research based on traditional models is relatively comprehensive, and some novel and effective multi-step or joint methods are proposed to extract multiple triples~\cite{li2019entity,wei2020novel,yu2020jointer,xie2021revisiting}.
For example, \rere~\cite{xie2021revisiting} carefully compares different types of multi-step settings and shows that the \emph{relation-then-entity} extraction paradigm exhibits a good performance since it suffers less from the problem of data imbalance, which is often encountered in relational triple extraction tasks. However, these methods cannot fully solve complex relational triple extraction tasks. Inspired by this, we propose to design an evaluation model and integrate this small model as a plug-in within the extraction framework based on LLMs.
% In conclusion, these methods cannot fully solve complex relational triples extraction tasks, and post-filtering is a general and available method to enhance the extraction results.

\begin{figure*}[!ht]
\begin{center}
\includegraphics[width=0.9\linewidth]{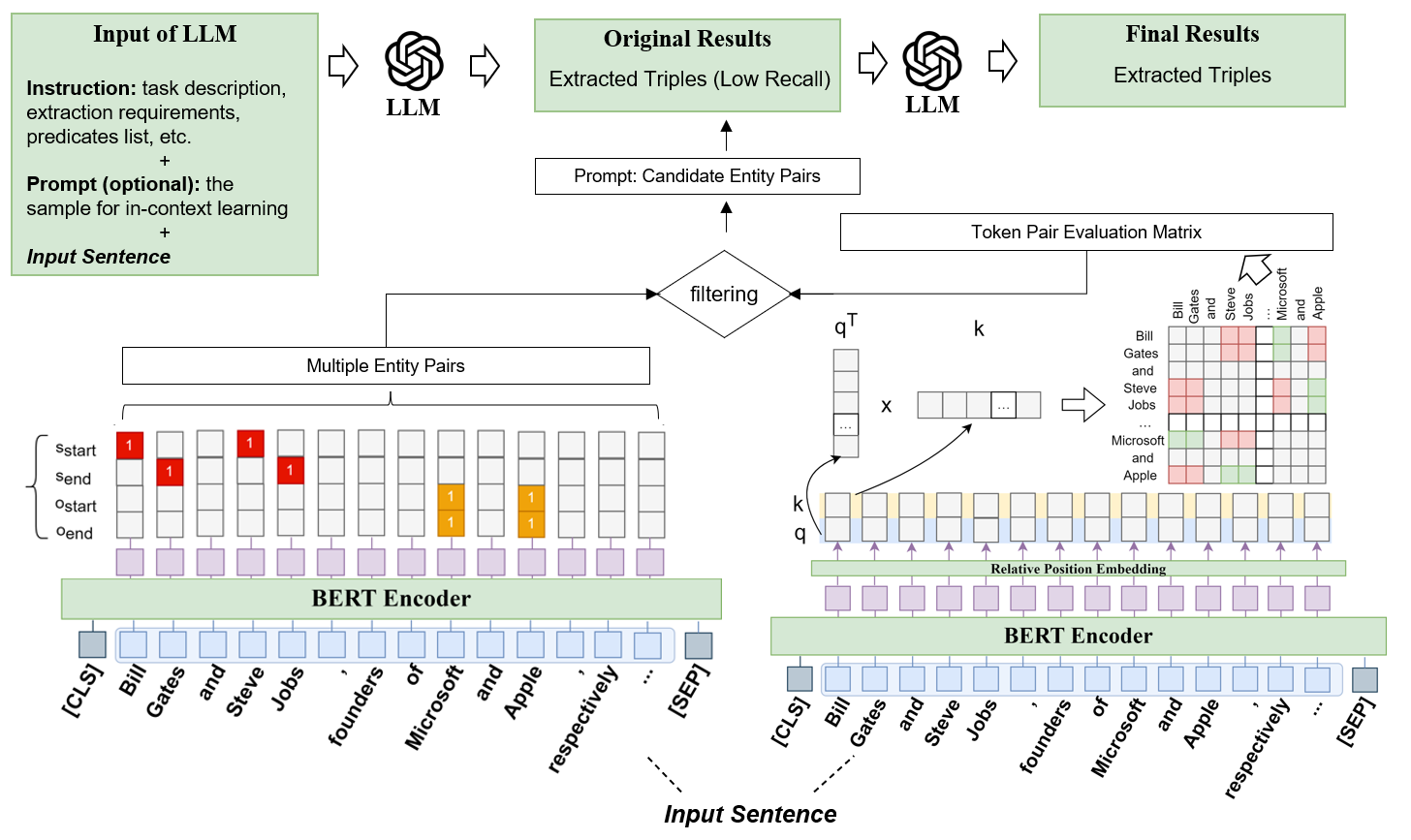} 
\caption{Model framework. On the bottom left is an arbitrary entity-extraction model. On the bottom right is our evaluation model, which outputs a token pair scoring matrix.}
\label{fig:mainmodel}
\end{center}
\end{figure*}

\section{Methods}

\subsection{Solution Framework}
\begin{figure}[!ht]
\begin{center}
\includegraphics[width=1\linewidth]{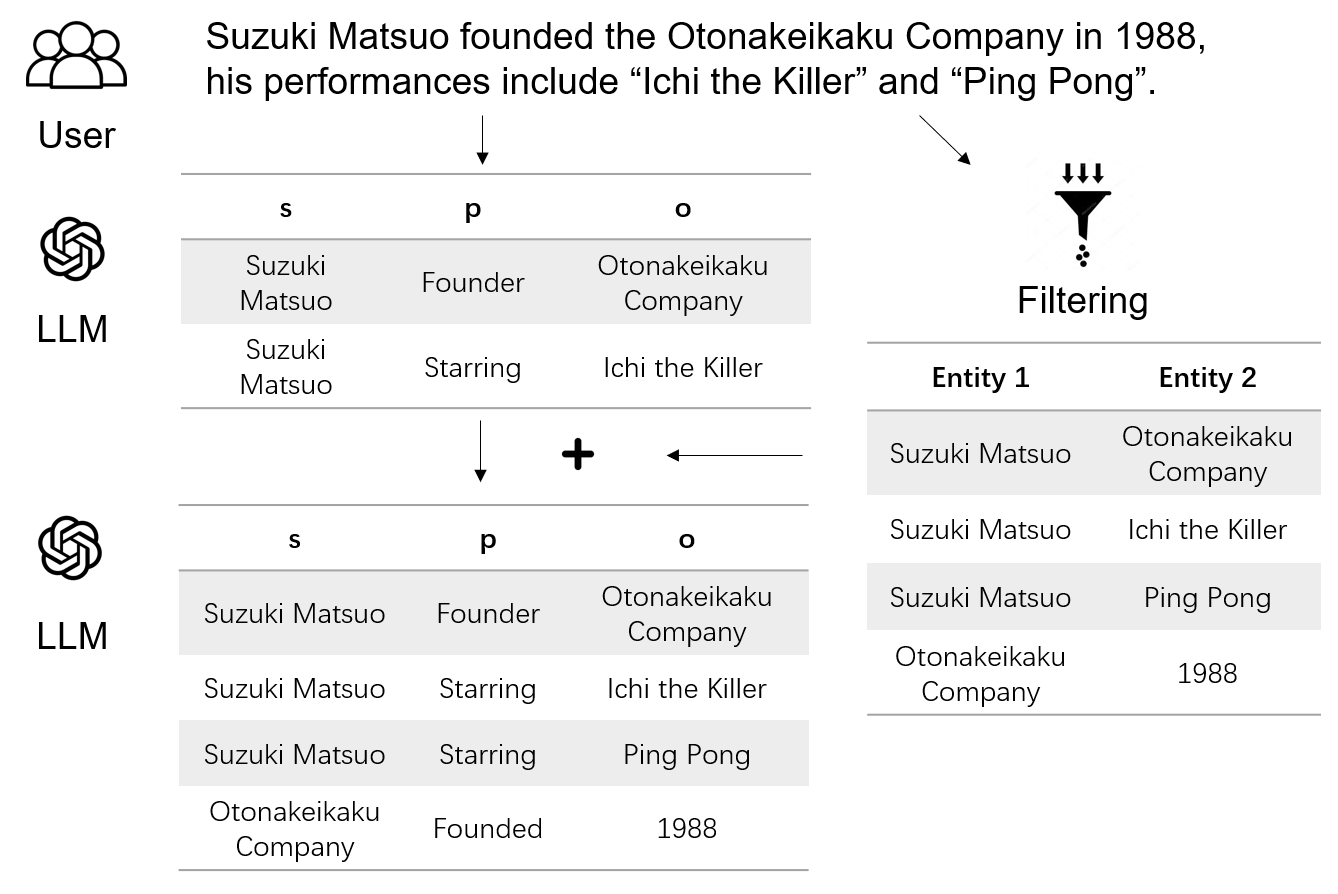} 
\caption{An example of the workflow of our Evaluation-Filtering method.}
\label{fig:demo}
\end{center}
\end{figure}

% The relational triple extraction task aims at extracting a set of triples from an input sentence, with an \emph{extraction model}.
Our LLM-based relational triple extraction framework comprises \textbf{two stages} (see Figure~\ref{fig:mainmodel}). In the first stage, the LLMs directly extract triples from sentences according to the provided instructions. Subsequently, in the second stage, we design an "evaluation-filtering" method, which extracts the positive entity pairs by our evaluation model and uses prompts to inform the LLMs that "these entity pairs may have certain relations in the relations list". These candidate pairs will be provided to the LLMs along with the instructions and the first-stage extraction results. LLMs will further scrutinize these candidates and assign appropriate relations based on their language comprehension capabilities, thereby achieving comprehensive and accurate extraction results. An example of the whole workflow is shown in Figure~\ref{fig:demo}.

\subsection{Basic Idea of Evaluation Model}
The \emph{evaluation model} (bottom-right part of Figure~\ref{fig:mainmodel}) uses a sentence (a list containing $N$ tokens) as the input and outputs a token pair evaluation matrix ($N*N$).
Each element in the matrix is an evaluation score for a token pair.
The evaluation score of a token pair $t_i$ and $t_j$ is used to compose the evaluation score of an entity pair $(s, o)$, where $s$ contains $t_i$ and $o$ contains $t_j$.
%$F(s,p,o)=F(s,o)=\hat{F}(t_i,t_j)$
Obviously, for an input sentence, no matter how many candidate pairs are to be evaluated, only one inference is needed to get the evaluation matrix. Our goal is to build such a model that scores candidate entity pairs based on the sentence from which the triples are extracted, as Problem 1 shows.

Clearly, this evaluation model could be used as a filter, removing the extracted candidate entity pairs with low scores while retaining those with high scores. After this filtering process, we can obtain a set of precise and complete positive samples (i.e., truly related entity pairs), which can then be supplied to LLMs as prompts to facilitate high-precision extraction of multiple relational triples.

\begin{problem}[Evaluation of entity pairs] 
Given a sentence $T$ and a candidate entity pair set $C$, the evaluation model outputs a score $F({s,o})$ for each pair $(s,o) \in C$.
\end{problem}

Moreover, in order to overcome the noisy entity problem, we use token-level representation to support any possible entities in the sentence.

\paragraph*{Rationality for providing entity pairs}
Note that we only evaluate entity pairs $(s, o)$ and provide them to LLMs, ignoring the predicate $p$.
The rationality of ignoring the predicate is as follows.
First, we find that in most real datasets, the entity pairs are more accurate than predicates in a labeled sample\footnote{E.g., in the NYT11 training set, there are 20\% wrong triples, but only 6\% wrong entity pairs.}. Second, the model structure based on entity pairs evaluation is more straightforward. It only needs to generate one evaluation matrix for a sentence, while the evaluation model for entire $(s, p, o)$ triples requires to generate k matrices, where k represents the number of relations contained within the sentence.
%%Thus, evaluation in terms of entity pairs is more reasonable than that with the predicate.
% Second, our experimental results show that 
% evaluation just by entity pair is already effective enough for the filtering of wrong triple candidates. Moreover, 
% providing these candidate entity pairs is adequate for LLMs to correctly extract missing triples.

\paragraph*{Rationality of token based representation}
Our model aims at evaluating any candidate entity pairs in the sentences.
However, extracting the entity span accurately is still a problem~\cite{dixit2019span,ji2020span}.
%, and the evaluation model does not have any prior entity information for the sentence.
For example, ``\textit{Gates and Steve}'' might be wrongly identified as an entity (in Figure~\ref{fig:labeling}).
Thus, it is necessary to evaluate the candidate "entities" represented by arbitrary tokens.

\subsection{Self Labeling}
The evaluation model has to distinguish between correct entity pairs and wrong entity pairs, which is a binary classification task, thus we need positive and negative training samples.
In original extraction datasets, a sentence is labeled with some triples, which correspond to some entity pairs with one of the target relations.
Obviously, these entity pairs are positive samples ($y=1$).
Then, it is important to obtain negative samples.
We generate negative samples with the following assumption:

\begin{assumption}
If a labeled sentence contains multiple triples, which involve multiple entities, then the unlabeled entity pairs are negative samples.
\label{assumption}
\end{assumption}

% As shown in Figure~\ref{fig:labeling}, for this sentence, most people will label two and only two triples \textit{(Microsoft, founderBy, Bill Gates)} and \textit{(Apple, founderBy, Steve Jobs)}, which involve four entities.
% This means that the sentence does not mention the relations between other entity pairs such as \textit{(Microsoft, Steve Jobs), (Microsoft, Apple)}, and \textit{(Bill Gates, Steve Jobs).}
% Thus, these entity pairs are labeled as negative samples ($y=-1$), which indicates that these entity pairs with any target relations should not be extracted in the sentence.

%\Red{
%(1) why works
    %previous works only consider what can be extracted
%    experimental: previous methods extract triples that are wrong, most of which are wrong entity-pairs (?)
%    theoretical: jiangjie LM doesn't know NOT (controllable?) so is extraction model
%(2) why exclude relation
%    (1) most of which are wrong entity-pair;
%    (2) collect negative triples require external semantics? 
%}

\paragraph*{Rationality of Assumption 1}
%We demonstrate the rationality of Assumption 1 from both theoretical and practical perspectives.
Assumption 1 will generate a false negative entity pair $(e_1, e_2)$ \emph{only} when all the four conditions are satisfied (``*'' means any relations or entities):
\begin{itemize}
\small
	\item $(e_1, *, e_2)$ is mentioned in the sentence.
	\item Any triples $(e_1, *, e_2)$ are not labeled in the sentence.
	\item Triple $(e_1, *, *)$ or $(*, *, e_1)$ is labeled in the sentence.
	\item Triple $(e_2, *, *)$ or $(*, *, e_2)$ is labeled in the sentence.
\end{itemize}
However, it is seldom the case that the four conditions are simultaneously met.
% No matter whether the triple extraction dataset comes from distant supervision or human annotation, 
The false negative case means that an annotator (no matter whether it is distant supervision via knowledge base or hand annotation via human) labels other triples in a sentence for both $e_1$ and $e_2$, but only misses the relation between them.

\begin{figure*}[!ht]
\begin{center}
\includegraphics[width=0.7\linewidth]{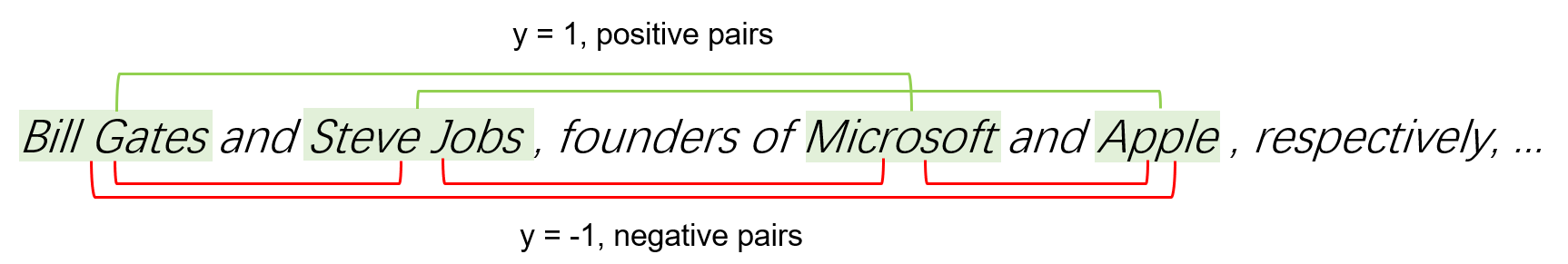} 
\caption{This sentence contains 6 entity pairs, but only 2 pairs are positive.}
\label{fig:labeling}
\end{center}
\end{figure*}

\paragraph*{Token-level labeling}
The above process generates entity pair samples, and the labeled token pairs can be simply and effectively obtained.
%but our evaluation model has to classify token pairs.
%The adaption process is simple and effective.
For example, in Figure~\ref{fig:labeling}, we simply split the negative entity pair \textit{(Microsoft, Steve Jobs)} into token pairs \textit{(Microsoft, Steve)} and \textit{(Microsoft, Jobs)}, which are labeled as negative token pairs.
This process not only increases the number of training pairs, but also enables the model to evaluate unseen or wrong entities, or even any token sequence.
%In other words, we don’t care about which target relation between these entity-pairs is, while they are all labeled as positive.
%This leads to a binary classifier, which has higher training effectiveness than a model with many relations.

Note that, for any other token pairs in the sentence (e.g. \textit{(founders, Microsoft)}), they are not labeled as negative or positive ($y=0$).
They will be masked in the training process of our evaluation model since we have no information about whether they are positive or negative.

% \paragraph*{Characteristics of the dataset}
% Under Assumption 1, we automatically construct a self-labeled evaluation dataset from the extraction dataset.
% %The self-labeled evaluation dataset has the following characteristics:
% %We argue that, for most popular triple extraction datasets, our self-labeled evaluation dataset is of high quality:
% We argue that, for most popular extraction datasets, our self-labeled evaluation dataset is relatively balanced, making the classification model easy to train.
% The ``balance'' here means that the number of positive samples and negative samples is close.
% Suppose a training sentence has $R$ triples that involve $E$ entities, there will be $R$ positive samples and $E^2-R$ negative samples.
% Thus, the ratio of the number of negative samples to the number of positive samples can be estimated by $\frac{E^2-R}{R} \approx O(E)$ (since the number of triples and entities are usually in the same order of magnitude, see Table~\ref{tab:complexstat}).
% Moreover, Table~\ref{tab:complexstat} also shows that, in many triple extraction datasets, the average number of entities ($E$) in a sentence is quite small\footnote{Although we actually use token pairs, the number of token pairs can be estimated by multiplying by the square of the average entity length, so the estimated ratio remains the same.}.

\subsection{Evaluation Model Structure}
Following the Transformer architecture, our evaluation model adopts a BERT-based encoder and an attention-like 2-dim decoder (as shown in the right part of Figure~\ref{fig:mainmodel}).

\subsubsection{Encoder}
%Due to the success of Transformer-based~\cite{vaswani2017attention} pre-trained language models, 
We use a regular Transformer model as our encoder.
Specifically, we use BERT~\cite{devlin2018bert} for English and RoBERTa~\cite{liu2019roberta} for Chinese.
They have the same network structure.

More formally, for an input sentence with $N$ tokens $[t_1, t_2, ..., t_N]$, where $t_1 = [CLS]$ and $t_N = [SEP]$ are fixed special tokens, the BERT encoder converts these tokens into hidden vectors $[\mathbf{h}_1, \mathbf{h}_2, ..., \mathbf{h}_N]$, where each $\mathbf{h}_i$ is a $d_1$-dimension vector.
In the BERT-base structure, $d_1=768$.

\subsubsection{2-dim Decoder}
For an input sentence with $N$ tokens, the BERT-based encoder encodes the tokens into $N$ vectors $\mathbf{[h_1, h_2, ..., h_N]}$.
Then, following the computation of the attention matrix in Transformers, we use a one-head self-attention to compute the 2-dim attention matrix as the output of the decoder.
In detail, we first use two linear layers to convert the vectors $\mathbf{h}_i$ to $d_2$-dimension vectors $\mathbf{q}_i$ and $\mathbf{k}_i$:
%, defined in Equation~\eqref{eq:qkline}:
\begin{equation}
\small
	\label{eq:qkline}
	\begin{aligned}
		\mathbf{q}_i = \mathbf{W}^{(q)}\mathbf{h}_i + \mathbf{b}^q, \\
		\mathbf{k}_i = \mathbf{W}^{(k)}\mathbf{h}_i + \mathbf{b}^k,
	\end{aligned}
\end{equation}
where $\mathbf{W}$ and $\mathbf{b}$ are trainable parameters of the linear layers, and $d_2=64$.
Then, we compute their scaled dot-product attention as the output:
%using Equation~\eqref{eq:dec}:
\begin{equation}
\small
	\label{eq:dec}
	A_{ij} = \mathbf{q}^T_i\mathbf{k}_j / \sqrt{d_2}.
\end{equation}

As proposed by Roformer~\cite{su2021roformer}, it is advantageous to add relative position embeddings (RoPE) before computing the attention output. 
The relative position embeddings $\mathbf{R}_i$ are realized by constructing sine and cosine functions that satisfy $\mathbf{R}_i^T\mathbf{R}_j=\mathbf{R}_{j-i}$, we refer the readers to ~\cite{su2021roformer} for technical details.
The intuition is that when encoding positional information $R_i$ and $R_j$ at position $i$ and $j$, the output attention will naturally contain the relative positional information.
The final form of the attention output is:
%defined by Equation~\eqref{eq:decpos}:

\begin{equation}
\small
	\label{eq:decpos}
	\begin{aligned}
		A'_{ij} &= (\mathbf{R}_i\mathbf{q}_i)^T(\mathbf{R}_j\mathbf{k}_j) / \sqrt{d_2} \\
		&=\mathbf{q}^T_i\mathbf{R}_{j-i}\mathbf{k}_j / \sqrt{d_2} .
	\end{aligned}
\end{equation}

Recall that this decoder (without the relative position embeddings) is only a part of regular one-head self-attention, although it has $O(N^2)$ outputs, its computational cost is smaller than the Transformer-based encoder.
Hence, the cost of training such an evaluation model is lower than a Transformer-based extraction model.

\subsection{Loss Function}
Since our task is a classification task with positive and negative labels, we use the binary cross-entropy loss function to train our evaluation model:
\begin{equation}
\small
	L = - \sum_{i,j: y_{ij} = 1}\log{(\sigma{(A'_{ij}))}} - \sum_{i,j: y_{ij} = -1}\log{(1-\sigma{(A'_{ij}))}},
\end{equation}
where $y_{ij}$ is the label of the token pair $(t_i, t_j)$, and $\sigma$ is the sigmoid function.
Note that, our task is not a pure binary classification task, since there are many unlabeled token pairs in our task.
Thus, the positive and negative examples are not complementary.
In the implementation of the loss, we ignore the part of unlabeled pairs (i.e. $y=0$).

\subsection{Candidate Pairs Evaluation}
After training such an evaluation model, we adapt the model to an existing extraction method to obtain better extraction results.
We score each candidate extracted pairs $(s, o)$, 
where $s=[t_{s_{st}}, t_{s_{st}+1}, ..., t_{s_{ed}}]$ and $o=[t_{o_{st}}, t_{o_{st}+1}, ..., t_{o_{ed}}]$ are sub-token-sequences in the given sentence.
Recall that, our evaluation model outputs a token pair evaluation matrix $A'$.
Based on this matrix, we compute the score between $s$ and $o$ by the mean of the matching scores of their tokens:
\begin{equation}
\small
	F(s, o) = \frac{\sum_{k=s_{st}}^{s_{ed}}\sum_{l=o_{st}}^{o_{ed}}A'_{kl}}{(s_{ed} - s_{st} + 1)(o_{ed} - o_{st} + 1)}
\end{equation}
where $s_{st}$ and $o_{st}$ are the indexes of the first element of token lists $s$ and $o$, $s_{ed}$ and $o_{ed}$ are the indexes of the last element of $s$ and $o$.

Finally, only $(s, o)$ satisfying $F(s, o) > 0$ will be added to the result.
Note that, we only need to predict the matrix $A'$ once for each sentence, no matter how many triples of this sentence should be evaluated, thus the evaluation process is efficient.

\begin{table*}[!htb]
	\centering
	\resizebox{2\columnwidth}{!}{
	\begin{tabular}{cccccccccccccccc}
		\hline
		\multirow{2}{*}{Dataset} &
		\multicolumn{3}{c}{$|T|$\textgreater{}=0} &
		\multicolumn{3}{c}{$|T|$\textgreater{}=30} &
		\multicolumn{3}{c}{$|T|$\textgreater{}=50} &
		\multicolumn{3}{c}{$|T|$\textgreater{}=70} &
		\multicolumn{3}{c}{$|T|$\textgreater{}=100} \\ \cline{2-16} 
		& \multicolumn{1}{c}{avgE} & avgR & \#sen & \multicolumn{1}{c}{avgE} & avgR & \#sen & \multicolumn{1}{c}{avgE} & avgR & \#sen &\multicolumn{1}{c}{avgE} & avgR & \#sen &\multicolumn{1}{c}{avgE} & avgR & \#sen \\ \hline

        NYT10 & \multicolumn{1}{c}{2.2}  & 1.7 & 5000 & \multicolumn{1}{c}{2.2}  & 1.8 & 4091 & \multicolumn{1}{c}{2.2}  & 1.8 & 1798 &\multicolumn{1}{c}{2.3}    & 1.9 & 441  &\multicolumn{1}{c}{2.3}    & 2.1 & 51  \\ 
   
  	NYT11-HRL & \multicolumn{1}{c}{2.0}  & 1.0 & 369 &  \multicolumn{1}{c}{2.0}  & 1.0 & 283 & \multicolumn{1}{c}{2.0}  & 1.0 & 120 & \multicolumn{1}{c}{2.0}    & 1.0 & 28 & \multicolumn{1}{c}{2.0}    & 1.0 & 3 \\ 

		SKE21     & \multicolumn{1}{c}{3.3}  & 2.4  & 1150 & \multicolumn{1}{c}{3.5}  & 2.6  & 901 &\multicolumn{1}{c}{3.9}  & 3.0  & 423 &\multicolumn{1}{c}{3.9}  & 3.0  & 202 &\multicolumn{1}{c}{4.0}  & 3.2  & 80 \\ 

        WikiKBP & \multicolumn{1}{c}{2.1}  & 1.1 & 182 & \multicolumn{1}{c}{2.2}  & 1.2 & 98 & \multicolumn{1}{c}{2.1}  & 1.0 & 25 &\multicolumn{1}{c}{2.3}    & 1.2 & 6  &\multicolumn{1}{c}{-}    & - & - \\ 
  
		\hline
  
	\end{tabular}
	}
	\resizebox{2\columnwidth}{!}{
	\begin{tabular}{cccccccccccccccc}
	    \hline
	     \multirow{2}{*}{Dataset} &
		\multicolumn{3}{c}{$|T|$\textgreater{}=50} &
		\multicolumn{3}{c}{$|T|$\textgreater{}=100} &
		\multicolumn{3}{c}{$|T|$\textgreater{}=150} &
		\multicolumn{3}{c}{$|T|$\textgreater{}=200} &
		\multicolumn{3}{c}{$|T|$\textgreater{}=250} \\ \cline{2-16} 
		& \multicolumn{1}{c}{avgE} & avgR & \#sen & \multicolumn{1}{c}{avgE} & avgR & \#sen & \multicolumn{1}{c}{avgE} & avgR & \#sen &\multicolumn{1}{c}{avgE} & avgR & \#sen &\multicolumn{1}{c}{avgE} & avgR & \#sen \\ \hline
		
		HacRED    & \multicolumn{1}{c}{7.1}  & 7.4  & 1500 &\multicolumn{1}{c}{7.4}  & 7.7  & 1372 &\multicolumn{1}{c}{8.2}  & 8.8  & 1012 &\multicolumn{1}{c}{9.1}  & 10.0 & 693 &\multicolumn{1}{c}{10.2}  & 11.4 & 410\\ \hline
	\end{tabular}
	}
	\caption{The statistics of complex sentences of testing datasets. $|T|$ means the number of tokens in the sentences.
		$|T| >= x$ only reports results for sentences with at least $x$ tokens.
		$avgE$, $avgR$ denote the average numbers of labeled entities, labeled triples in the sentence, respectively.
        $\#sen$ denotes the number of sentences.}
	\label{tab:complexstat}
\end{table*}

\subsection{Parameter-Efficient Fine-Tuning for LLMs}
For multiple relational triple extractions, employing instruction-tuning or in-context learning (ICL) to guide LLMs, as is done for general tasks, often yields unsatisfactory results. This is because LLMs possess strong generalization capabilities and language comprehension, leading them to inexactly recognize the span of entities or relations. For instance, they may extract predicates not presented in the predicate list, or consider book titles as part of an entity, even when their extraction range is explicitly limited through prompts. Consequently, parameter fine-tuning is necessary to adapt the model to the corresponding datasets and to potentially non-natural language representations of predicates (e.g. NYT10 dataset).

In this paper, we mainly adopt the LoRA technology ~\cite{hu2021lora}. LoRA is a parameter-efficient fine-tuning (PEFT) method. It freezes the large-scale parameters of a pre-trained model and simulates parameter changes through low-rank decomposition of the matrix, thereby adapting the model to downstream tasks with small-scale parameter adjustments. Compared to full-parameter fine-tuning, this method is more time-efficient and requires less computing resources and storage. 
% and improves performance in relational triple extraction tasks compared to the origin model.

% It has been observed that direct methods such as instruction-tuning or in-context learning (ICL) often yield unsatisfactory results in this task, as demonstrated in Section 4. The model tends to extract some relations that are not listed, even when its extraction range is explicitly limited by prompts. Furthermore, the recognition of entities often does not match the rules of the dataset itself, such as the addition of quotation marks around book titles.

% However, we cannot directly judge these extraction results as errors. Therefore, parameter fine-tuning is required to adapt the model to the corresponding dataset and to potentially non-natural language representations of relations.

\subsection{Instruction Template}
To better guide LLMs in performing multiple relational triple extraction tasks, we design an instruction template that explicitly includes the task description, the restricted range of extracted predicates, the output format, and other requirements. We also explicitly require the model to extract as many relation triples as possible. Additionally, for the original large model without PEFT, a complete input-output example can also be placed after the instruction template for in-context learning.
% The instruction template is as follows:

% \paragraph*{Instruction Template}
% Predefine the following relation list: \textit{predicates\_list}, please extract all triples containing the above relations from the following sentences.
% Note that the relation name of the triple must be selected from the above list, and other relations not listed are not considered. Please output according to the specified format below:
% \textit{[{"s": subject1, "o": object1, "p": relation1}, {"s": subject2, "o": object2, "p": relation2},...]}. Note that the triple may not only have two, please imitate this format and output all triples that meet the requirements.
% Again, it is emphasized that the relation of the triples output must be selected from the predefined list above, and no relation not in the list can be output. At the same time, please output as many triples as possible that meet the requirements.
% Please according to the input, output all triples containing the above relationship according to the format requirements. Note that when the entity (subject or object) can be split into two words (such as a comma or comma in the middle), it should be split into two triples instead of merging into one triple.

After the evaluation model extracts candidate entity pairs, these candidates will be provided to the LLMs as part of the prompt, along with the aforementioned instructions and the first-stage extraction results, to guide the model in completing the extraction. See Appendix~\ref{sec:template} for detailed prompts.
% TODO: 我们的prompt模板见附录B

\section{Experiments}
\subsection{Datasets}
We evaluate our methods on several public and downloadable complex extraction datasets, including NYT series~\cite{riedel2010modeling, takanobu2019hierarchical}, Wiki-KBP~\cite{ling2012fine}, SKE21~\cite{xie2021revisiting} and HacRED~\cite{hacred}, which are challenging for many extraction methods.
Table~\ref{tab:complexstat} shows their statistics and shows that sentences in these datasets often have more than one triple. A brief introduction to these datasets is provided in Appendix~\ref{sec:dataset}.

For the NYT series, the relations in the original data are labeled in structured formats (e.g. \textit{/location/location/contains}), which is not easily comprehensible to large models. Therefore, before conducting experiments, we converted these relations into natural language with similar meanings(e.g. \textit{location contains}) to enhance the model's understanding.

\subsection{Comparing Methods and Metrics}
We apply our method to some recently popular LLMs such as Qwen-7B, Vicuna-13B, and Llama-13B, and compare the triple extraction performance of our framework including evaluation-filtering with the base large models. To better assess the effectiveness of our method in complex scenarios, we have also specifically calculated the metrics in cases involving multiple triples.

Additionally, we also apply our methods to pretrained-language-model-based approaches including CasRel~\cite{wei2020novel}, TPLinker~\cite{wang2020tplinker}, and \rere~\cite{xie2021revisiting}, to examine the effectiveness of the evaluation model and evaluate its improvement over small models.

We report standard precision (Prec.), recall (Reca.), and F1 scores for all the experiments.
We mainly focus on the \textbf{exact match result}, which is also the main consideration of current extraction methods.
Note that, some triples extracted by the model may be deemed errors when calculating metrics due to the synonyms or the addition of certain symbols. These could also be considered as correct results by manual evaluation. However, employing exact matching makes the evaluation and comparison of results more rigorous and credible.
% Note that, some early methods use partial match on these datasets, which cannot fully reflect the true effectiveness in real-world applications. 

\begin{table*}[!htb]
	\centering
	\resizebox{2.05\columnwidth}{!}{
		\begin{tabular}{p{3.8cm}rrrrrrrrrrrrrrr}
			\toprule
			& \multicolumn{3}{c}{SKE21}  & \multicolumn{3}{c}{HacRED}  & \multicolumn{3}{c}{NYT10}  & \multicolumn{3}{c}{NYT11-HRL}  & \multicolumn{3}{c}{WikiKBP}     \\
			\cmidrule{2-16}       & \multicolumn{1}{c}{Prec.} & \multicolumn{1}{c}{Reca.} & \multicolumn{1}{c}{F1} & \multicolumn{1}{c}{Prec.} & \multicolumn{1}{c}{Reca.} & \multicolumn{1}{c}{F1} & \multicolumn{1}{c}{Prec.} & \multicolumn{1}{c}{Reca.} & \multicolumn{1}{c}{F1} & \multicolumn{1}{c}{Prec.} & \multicolumn{1}{c}{Reca.} & \multicolumn{1}{c}{F1} & \multicolumn{1}{c}{Prec.} & \multicolumn{1}{c}{Reca.} & \multicolumn{1}{c}{F1} \\
			\midrule
			Qwen-7b (w/o peft)& 17.59 & 26.18 & 21.04 & 5.72 & 6.10 & 5.90 & 8.18 & 6.12 & 7.00 & 8.08 & 16.19 & 10.78 & 6.64 & 19.50 & 9.91  \\
            \text{ } \text{ } +Ours & 42.21 & 40.14 & \textbf{41.15} & 19.56 & 15.06 & \textbf{17.02} & 10.07 & 6.75 & \textbf{8.08} & 15.87 & 24.13 & \textbf{19.14} & 9.02 & 18.00 & \textbf{12.02} \\           
            \text{ } \text{ } \#Triples\textgreater{}=$t$ & 31.53 & 29.77 & 30.63 & 6.04 & 4.37 & 5.07 & 12.89 & 2.96 & 4.81 & 5.63 & 10.67 & 7.37 & 7.46 & 12.82 & 9.43 \\
            \text{ } \text{ } +Ours, \#Triples\textgreater{}=$t$ & 54.62 & 39.10 & \textbf{45.57} & 21.43 & 13.98 & \textbf{16.93} & 13.13 & 5.20 & \textbf{7.45} & 12.50 & 16.67 & \textbf{14.29} & 8.47 & 12.82 & \textbf{10.20} \\
    % &  &  & +\% &  &  & +\% &  &  & \% &  &  & +\% &  &  & +\% \\
    
    \midrule
			Qwen-7b (w/ peft)& 50.34 & 70.16 & 58.62 & 44.51 & 41.47 & 42.94 & 66.30 & 56.27 & 60.88 & 56.48 & 58.01 & 57.24 & 42.54 & 48.50 & 45.33  \\
            \text{ } \text{ } +Ours & 48.17 & 77.90 & \textbf{59.53} & 42.89 & 49.95 & \textbf{46.15} & 65.54 & 75.37 & \textbf{70.11} & 56.95 & 67.24 & \textbf{61.67} & 39.58 & 56.00 & \textbf{46.38} \\           
            \text{ } \text{ } \#Triples\textgreater{}=$t$ & 63.27 & 65.40 & 64.31 & 51.54 & 39.77 & 44.89 & 81.49 & 43.30 & 56.55 & 76.74 & 44.00 & 55.93 & 34.78 & 20.51 & 25.81 \\
            \text{ } \text{ } +Ours, \#Triples\textgreater{}=$t$ & 61.36 & 73.84 & \textbf{67.02} & 50.84 & 47.37 & \textbf{49.04} & 78.26 & 66.53 & \textbf{71.92} & 60.94 & 52.00 & \textbf{56.12} & 35.00 & 35.90 & \textbf{35.44} \\
    % &  &  & +\% &  &  & +\% &  &  & \% &  &  & +\% &  &  & +\% \\

   \midrule
			Llama-13b (w/ peft)& 33.65 & 24.05 & 28.05 & 12.89 & 9.28 & 10.79 & 18.05 & 54.90 & 27.17 & 13.85 & 52.93 & 21.95 & 18.84 & 76.00 & 30.19  \\
            \text{ } \text{ } +Ours & 30.16 & 36.06 & \textbf{32.84} & 27.98 & 15.52 & \textbf{19.96} & 19.04 & 49.78 & \textbf{27.54} & 24.13 & 65.92 & \textbf{35.32} & 20.96 & 76.00 & \textbf{32.86} \\           
            \text{ } \text{ } \#Triples\textgreater{}=$t$ & 37.08 & 21.35 & 27.10 & 17.03 & 8.33 & 11.19 & 36.68 & 47.60 & 41.43 & 38.83 & 53.33 & 44.94 & 20.22 & 46.16 & 28.12 \\
            \text{ } \text{ } +Ours, \#Triples\textgreater{}=$t$ & 44.50 & 29.17 & \textbf{35.24} & 31.35 & 13.67 & \textbf{19.04} & 51.38 & 56.35 & \textbf{53.75} & 59.09 & 52.00 & \textbf{55.32} & 32.00 & 61.54 & \textbf{42.10} \\
    % &  &  & +\% &  &  & +\% &  &  & \% &  &  & +\% &  &  & +\% \\

  \midrule
			Vicuna-13b (w/ peft)& 69.30 & 51.75 & 59.25 & 34.36 & 33.35 & 33.85 & 71.02 & 62.89 & \textbf{66.71} & 32.92 & 56.36 & 41.57 & 17.06 & 18.00 & 17.52  \\
            \text{ } \text{ } +Ours & 68.20 & 67.96 & \textbf{68.08} & 53.49 & 40.88 & \textbf{46.34} & 60.98 & 65.86 & 63.33 & 34.20 & 61.37 & \textbf{43.93} & 37.45 & 47.00 & \textbf{41.69} \\           
            \text{ } \text{ } \#Triples\textgreater{}=$t$ & 84.55 & 29.62 & 43.88 & 45.86 & 30.97 & 36.97 & 76.29 & 43.53 & 55.43 & 55.41 & 54.67 & 55.03 & 30.00 & 23.08 & 26.09 \\
            \text{ } \text{ } +Ours, \#Triples\textgreater{}=$t$ & 74.59 & 60.90 & \textbf{67.05} & 61.98 & 38.68 & \textbf{47.63} & 70.40 & 51.76 & \textbf{59.66} & 59.49 & 62.67 & \textbf{61.04} & 41.38 & 30.77 & \textbf{35.29} \\
    % &  &  & +\% &  &  & +\% &  &  & \% &  &  & +\% &  &  & +\% \\
			\bottomrule
		\end{tabular}%
		% \addtolength{\tabcolsep}{2.5pt}
	}
	\caption{The main extraction results. The "w/ peft" means that parameter-efficient fine-tuning (LoRA) of base LLMs is done before triples extracting, based on a part of the train set (about 800 sentences).
     Better exact match F1 scores are marked \textbf{bold}. The threshold $t$ is 2 for WikiKBP and NYT11-HRL since the most complex sentence only contains 4 triples in these datasets, and is 5 for other datasets.
     %CGT means the re-implemented version filter module in \cite{ye2021contrastive}
     %(w/o RoPE) means that the evaluation model does not use relative position embedding.
     }
	\label{tab:mainresult}%
\end{table*}%

\subsection{Effectiveness of Our Method}

%sentence length
The results in Table~\ref{tab:mainresult} demonstrate that with the assistance of our evaluation-filtering model, the triple extraction results of various LLMs on different Chinese and English datasets have been significantly enhanced. By using the filtered candidate pairs as prompts, compared to the basic LLMs, the recall rate in the multiple relational triple extractions task can be stably and significantly improved (more than 10\%) in most cases, with only a risk of slightly reduced precision. In fact, the precision of models that include evaluation-filtering will also be improved in many cases. 
% This is because the candidate pairs provided by our evaluation model have high precision (the ablation study results can verify this, see Table~\ref{tab:ablation}), so supplementing the missing triples for large models can also ensure the precision of the final results.

We specifically focused on relational triple extraction in more complex scenarios, setting a minimum number of triples for each dataset (2 for WikiKBP and NYT11-HRL, 5 for others), and only considering sentences containing more than this number of triples to assess the model's extraction effect on them. 
The results indicate that when a sentence contains a substantial number of triples, the direct application of LLMs to extract relational triples based on instructions often yields poor results, irrespective of whether fine-tuning on labeled data. Notably, the recall in most cases is significantly lower than the precision.
In contrast, on complex sentences (number of triples\textgreater{}=$t$) in various datasets, and the most complex dataset HacRED, our evaluation-filtering method can significantly enhance the recall of the extraction results, while also improving precision in most cases. 
% Furthermore, for complex sentences in the SKE21 dataset, our model has achieved a overwhelming doubling in the recall rate of the results.

\begin{figure}[!htb]
	\centering
	\begin{minipage}[t]{0.232\textwidth}
        \centering
        \includegraphics[width=\linewidth]{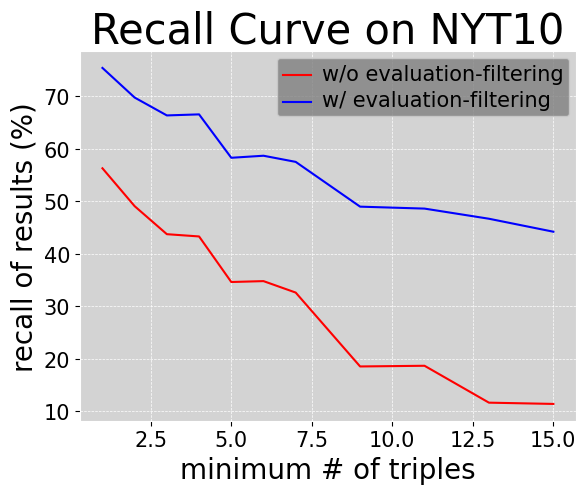}
    \end{minipage}
    \begin{minipage}[t]{0.232\textwidth}
        \centering
        \includegraphics[width=\linewidth]{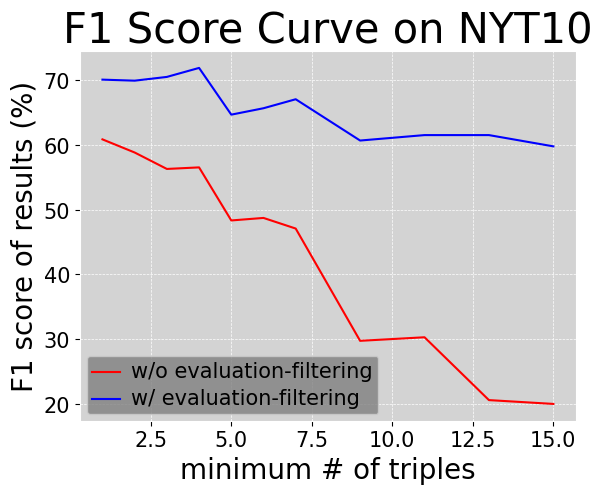}
    \end{minipage}
	\caption{Recall and F1-score curve of Qwen-7B (w/ peft) on NYT10, with and without our evaluation-filtering method. Minimum \# of triples means we only consider sentences that contain a number of triples greater than this value. Note that the coordinates do not start from 0.}
	\label{fig:ref_curve}
\end{figure}

In Figure~\ref{fig:ref_curve}, we use the NYT10 dataset, categorizing the test sentences based on the number of triples they contain, and utilize the Qwen-7B model to extract triples from sentences of different complexity levels. The results show that as the number of triples within a sentence increases, our model demonstrates a progressively noticeable improvement in the recall of relational triple extraction results, compared to the base model. Moreover, it can maintain the F1 score of the results at a relatively high level. This suggests that our method is particularly effective for extracting multiple relational triples from complex sentences, and it can sustain a high level of precision of results.

In addition, the results of the small extraction model in Table~\ref{tab:small} show that our method achieves a large precision improvement with a small recall decline, which leads to a better F1 score. This indicates that our evaluation model can accurately and reliably obtain candidate pairs, which can be applied to the traditional small extraction model to improve the performance of multiple relational triple extraction.

%In addition, our evaluation model improves different extraction methods. 
%As shown in Table~\ref{tab:mainen}, our evaluation model not only improves \rere, but also CasRel and TPLinker.

\begin{table*}[!htb]
	\centering
	\resizebox{1.8\columnwidth}{!}{
		\begin{tabular}{p{3.8cm}rrrrrrrrrrrr}
			\toprule
			& \multicolumn{3}{c}{NYT10}  & \multicolumn{3}{c}{NYT10-HRL}  & \multicolumn{3}{c}{SKE21}  & \multicolumn{3}{c}{HacRED}     \\
			\cmidrule{2-13}       & \multicolumn{1}{c}{Prec.} & \multicolumn{1}{c}{Reca.} & \multicolumn{1}{c}{F1} & \multicolumn{1}{c}{Prec.} & \multicolumn{1}{c}{Reca.} & \multicolumn{1}{c}{F1} & \multicolumn{1}{c}{Prec.} & \multicolumn{1}{c}{Reca.} & \multicolumn{1}{c}{F1} & \multicolumn{1}{c}{Prec.} & \multicolumn{1}{c}{Reca.} & \multicolumn{1}{c}{F1} \\
			\midrule
			TPLinker & 84.96 & 89.66 & 87.25 & 74.31 & 61.06 & 67.04 & 72.73 & 77.94 & 75.24 & 54.64 & 61.21 & 57.74  \\
	      \text{ } \text{ } + Ours & 86.87 & 89.36 & \textbf{88.10} & 74.79 & 60.86 & \textbf{67.11} & 81.31 & 77.63 & \textbf{79.43} & 61.78 & 59.04 & \textbf{60.38} \\
    \midrule
			CasRel & 83.82 & 87.63 & 85.69 & 70.25 & 65.11 & 67.58 & 84.24 & 67.50 & 74.95 & 62.62 & 34.62 & 44.59 \\
            \text{ } \text{ } + Ours & 88.23 & 87.40 & \textbf{87.81} & 72.35 & 64.88 & \textbf{68.41} & 84.89 & 67.42 & \textbf{75.15} & 69.48 & 34.18 & \textbf{45.82} \\
   \midrule
			\rere & 81.28 & 89.16 & 85.04 & 68.66 & 63.77 & 66.12 & 81.01 & 82.15 & 81.58 & 46.42 & 61.37 & 52.86 \\
			\text{ } \text{ }  + Ours & 87.03 & 88.80 & \textbf{87.90} & 71.32 & 63.61 & \textbf{67.42} & 83.44 & 81.68 & \textbf{82.55} & 69.92 & 59.37 & \textbf{64.21}\\
			\bottomrule
		\end{tabular}%
		% \addtolength{\tabcolsep}{2.5pt}
	}
	\caption{The main evaluation results of different small models.
     We only report the results for sentences with at least 50 tokens. 
     Best exact match F1 scores are marked \textbf{bold}. 
     %CGT means the re-implemented version filter module in \cite{ye2021contrastive}
     %(w/o RoPE) means that the evaluation model does not use relative position embedding.
     }
	\label{tab:small}%
\end{table*}%

\subsection{Further Analysis}
 % (Ablation study)
In Table~\ref{tab:ablation}, we try three ablation settings.
First, we remove the second stage of the framework, that is, the LLMs extraction part after receiving the candidate pairs prompt. 
Instead, we incorporated an additional relation classification model prior to the evaluation model. In other words, we use a small extraction model (\rere~\cite{xie2021revisiting}) to extract triples, which are then filtered according to the evaluation model. The filtered triples are combined with the first-stage LLMs' extraction results as the final results. The results indicate that the omission of the LLMs in the second stage leads to a decrease in the precision and F1 score of triple extraction results. Therefore, a large model in the second stage is still necessary for judgment and relation identification.
% Instead, we only filtered the initial extraction results of the LLMs, based on the token pair evaluation matrix obtained from the evaluation model. The results indicate that our evaluation model can accurately and reliably obtain candidate pairs. However, the recall of the results obtained directly through the first stage remains quite low, especially after filtering. 
% Therefore, a large model in the second stage is still necessary to supplement the results.

Second, we remove the first stage of the framework, that is, when inputting instructions and original sentences, we also provide the LLMs evaluation-filtering prompt, which will strictly limit the scope of triple extraction to the candidates provided by the evaluation model. The results show that our model can still enhance the recall rate of multiple triple extraction, but less effectively compared to the complete framework. This could be attributed to the presence of positive entity pairs that the evaluation model fails to recognize. However, without stringent restrictions, LLMs are capable of identifying and retaining these results.

Third, we remove the filtering step in the framework, that is, directly provide all entity pairs recognized by the entity-extraction model as prompts to the LLMs. The results show that the precision and F1 score of extraction results significantly decrease. This suggests that our evaluation-filtering method is indispensable.

\begin{table}[!htb]
	\centering
	\resizebox{1\columnwidth}{!}{
		\begin{tabular}{p{3.8cm}rrrrrr}
			\toprule
		Models (w/ peft)	& \multicolumn{3}{c}{SKE21 ($t$=7)} & \multicolumn{3}{c}{NYT11-HRL ($t$=3)}    \\
			\cmidrule{2-7}       & \multicolumn{1}{c}{Prec.} & \multicolumn{1}{c}{Reca.} & \multicolumn{1}{c}{F1} & \multicolumn{1}{c}{Prec.} & \multicolumn{1}{c}{Reca.} & \multicolumn{1}{c}{F1} \\
    \midrule
			Qwen-7b + Ours & 68.32 & 74.46 & \textbf{71.26} & 60.76 & 60.95 & \textbf{60.86} \\
            \text{ } \text{ } w/o stage 2 & 64.69 & 75.01 & 69.47 & 53.85 & 62.23 & 57.73 \\           
            \text{ } \text{ } w/o stage 1 & 72.77 & 68.86 & 70.76 & 63.33 & 50.67 & 56.30 \\
            \text{ } \text{ } w/o pairs filtering & 65.12 & 66.04 & 65.57 & 58.86 & 59.05 & 58.95 \\
    % &  &  & +\% &  &  & +\% &  &  & \% &  &  & +\% &  &  & +\% \\

   \midrule
			Llama-13b + Ours & 53.79 & 28.06 & 36.88 & 52.37 & 60.00 & 55.92 \\
            \text{ } \text{ } w/o stage 2 & 40.23 & 41.19 & \textbf{40.70} & 36.38 & 62.20 & 45.91 \\           
            \text{ } \text{ } w/o stage 1 & 63.11 & 25.06 & 35.87 & 68.11 & 52.00 & \textbf{58.97} \\
            \text{ } \text{ } w/o pairs filtering & 44.75 & 27.25 & 33.87 & 37.76 & 49.33 & 42.77 \\
    % &  &  & +\% &  &  & +\% &  &  & \% &  &  & +\% &  &  & +\% \\

  \midrule
			Vicuna-13b + Ours & 71.49 & 56.83 & \textbf{63.33} & 58.82 & 57.14 & \textbf{57.97} \\
            \text{ } \text{ } w/o stage 2 & 65.26 & 59.90 & 62.92 & 44.00 & 61.33 & 51.24  \\           
            \text{ } \text{ } w/o stage 1 & 92.76 & 36.06 & 51.93 & 70.91 & 41.67 & 52.49  \\
            \text{ } \text{ } w/o pairs filtering  & 63.97 & 57.67 & 60.66 & 32.38 & 55.18 & 40.82  \\
    % &  &  & +\% &  &  & +\% &  &  & \% &  &  & +\% &  &  & +\% \\
			\bottomrule
		\end{tabular}%
		% \addtolength{\tabcolsep}{2.5pt}
	}
	\caption{The ablation experiments results of different LLMs.
     We only report the results for sentences containing at least $t$ triples. Best exact match F1 scores are marked \textbf{bold}.
     %CGT means the re-implemented version filter module in \cite{ye2021contrastive}
     %(w/o RoPE) means that the evaluation model does not use relative position embedding.
     }
	\label{tab:ablation}%
\end{table}%

\section{Conclusion}
In this paper, we propose an evaluation model that can act as a filter to assess and identify entity pairs that have relations, thereby providing high-precision candidates for the subsequent extraction process. 

We incorporate this evaluation model into our proposed evaluation-filtering framework for LLMs multiple relation triple extraction. The candidates filtered by the evaluation model are integrated into the extraction process of LLMs in the form of prompts. This effectively addresses the issue of low recall rate in triple extraction tasks performed by LLMs, without diminishing precision. 

The experimental results that derived from multiple LLMs and datasets validate the effectiveness and completeness of our framework. Additionally, we confirm that our evaluation model can also be implemented in traditional small extraction models to enhance their precision and F1 score.

\section*{Acknowledgements}

This work was supported by Chinese NSF Youth Fund (No. 62102095), Major Research Plan (No. 92270121), and Shanghai Science and Technology Innovation Action Plan (No.21511100401).The computations in this research were performed using the CFFF platform of Fudan University.

\section*{Limitations}
% 复杂度讨论（可放在limitation里面说）

\paragraph*{Extraction Performance}
% 尽管我们的模型非常有效，但仍然有一定比例的三元组没有被抽取到最终结果中。一方面原因是，少量的related实体对并没有被评估模型正确赋值；另一方面则是由于大模型本身判断错误、或者没有正确地赋予关系。此外，第二阶段中大模型也很难完全删去第一阶段的错误三元组。后续研究可以探索对评估模型的优化、以及模型协同方法抽取准确度及召回率的进一步提高。
Despite the effectiveness of our model, the overall extraction results may still miss some correct triples and contain errors. On the one hand, a small amount of related entity pairs are not correctly evaluated by the evaluation model. On the other hand, it is difficult for LLMs to completely avoid errors or omissions in the second stage, although we prompt them to pick the correct candidate pairs and recheck the original results. Subsequent research could explore the optimization of the evaluation model, as well as further improvements in the extraction precision and recall of the model collaboration approach.

\paragraph*{Complexity of Our Method}
% 与基于提示引导大模型直接完成抽取任务相比，我们的方法在时间上消耗更久。时间上的开销主要源于xxxx（第二阶段大模型）。我们建议，xxxx。
% For the tradeoffs associated with using the proposed method, our method is indeed more complicated compared with the direct application of LLMs, but its inference time is only a change in the constant coefficient, rather than an increase in the exponential term. When dealing with complex sentences, to obtain stable effect improvement, our method is more suitable. Due to space constraints of the article, we have not added a detailed discussion of method complexity and application suggestions. This part will be added in the next version. Thank you for your suggestions.
Our framework involves multiple components and requires the LLMs to perform extraction twice. Our method is more complicated and more time-consuming with the direct application of LLMs, and its inference time roughly doubled. To obtain stable effect improvement when dealing with complex sentences, our method is more suitable, while for simple extraction tasks, we suggest single-stage direct extraction.

% \section*{Ethics Statement}
% \paragraph*{Use of Human Annotations}
% Human annotations are only used in methodological research at the beginning of the work, to assist in analyzing the feasibility of the proposed solution. Annotators consented to the use of data for research purposes. We ensure that the privacy of all annotators is protected throughout the annotation process, and all of them are adequately paid according to local standards. Human annotations are not applied during the evaluation of our method.
% \paragraph*{Risks}
% In this paper, all datasets are obtained from public sources. The datasets adopted have been anonymized and do not contain offensive information. However, we cannot guarantee that the datasets do not contain socially harmful or toxic language.

% \section{Providing References}

% \subsection{Bibliographical References} 

% Bibliographical references should be listed in alphabetical order at the end of the paper. The title of the section, ``Bibliographical References'', should be a Level 1 Heading. The first line of each bibliographical reference should be justified to the left of the column, and the rest of the entry should be indented by 0.35 cm.

% The examples provided in Section~\ref{sec:reference} (some of which are fictitious references) illustrate the basic format required for papers in conference proceedings, books, journal articles, PhD theses, and books chapters.

\nocite{*}
\section*{Bibliographical References}\label{sec:reference}

\bibliographystyle{lrec-coling2024-natbib}
\bibliography{lrec-coling2024-example}

\begin{thebibliography}{42}
\expandafter\ifx\csname natexlab\endcsname\relax\def\natexlab#1{#1}\fi

\bibitem[{Agrawal et~al.(2022)Agrawal, Hegselmann, Lang, Kim, and
  Sontag}]{agrawal2022large}
Monica Agrawal, Stefan Hegselmann, Hunter Lang, Yoon Kim, and David Sontag.
  2022.
\newblock Large language models are zero-shot clinical information extractors.
\newblock \emph{arXiv preprint arXiv:2205.12689}.

\bibitem[{Allingham et~al.(2023)Allingham, Ren, Dusenberry, Gu, Cui, Tran, Liu,
  and Lakshminarayanan}]{pmlr-v202-allingham23a}
James~Urquhart Allingham, Jie Ren, Michael~W Dusenberry, Xiuye Gu, Yin Cui,
  Dustin Tran, Jeremiah~Zhe Liu, and Balaji Lakshminarayanan. 2023.
\newblock \href {https://proceedings.mlr.press/v202/allingham23a.html} {A
  simple zero-shot prompt weighting technique to improve prompt ensembling in
  text-image models}.
\newblock In \emph{Proceedings of the 40th International Conference on Machine
  Learning}, volume 202 of \emph{Proceedings of Machine Learning Research},
  pages 547--568. PMLR.

\bibitem[{Bai et~al.(2023)Bai, Bai, Chu, Cui, Dang, Deng, Fan, Ge, Han, Huang
  et~al.}]{bai2023qwen}
Jinze Bai, Shuai Bai, Yunfei Chu, Zeyu Cui, Kai Dang, Xiaodong Deng, Yang Fan,
  Wenbin Ge, Yu~Han, Fei Huang, et~al. 2023.
\newblock Qwen technical report.
\newblock \emph{arXiv preprint arXiv:2309.16609}.

\bibitem[{Brown et~al.(2020)Brown, Mann, Ryder, Subbiah, Kaplan, Dhariwal,
  Neelakantan, Shyam, Sastry, Askell et~al.}]{brown2020language}
Tom Brown, Benjamin Mann, Nick Ryder, Melanie Subbiah, Jared~D Kaplan, Prafulla
  Dhariwal, Arvind Neelakantan, Pranav Shyam, Girish Sastry, Amanda Askell,
  et~al. 2020.
\newblock Language models are few-shot learners.
\newblock \emph{Advances in neural information processing systems},
  33:1877--1901.

\bibitem[{Cheng et~al.(2021)Cheng, Liu, Qu, Zhao, Liang, Wang, Huai, Yuan, and
  Xiao}]{hacred}
Qiao Cheng, Juntao Liu, Xiaoye Qu, Jin Zhao, Jiaqing Liang, Zhefeng Wang,
  Baoxing Huai, Nicholas~Jing Yuan, and Yanghua Xiao. 2021.
\newblock \href {https://doi.org/10.18653/v1/2021.findings-acl.249}
  {{H}ac{RED}: A large-scale relation extraction dataset toward hard cases in
  practical applications}.
\newblock In \emph{Findings of the Association for Computational Linguistics:
  ACL-IJCNLP 2021}, pages 2819--2831, Online. Association for Computational
  Linguistics.

\bibitem[{Chu et~al.(2020)Chu, Jiang, Xiao, and Wang}]{chu2020insrl}
Zhendong Chu, Haiyun Jiang, Yanghua Xiao, and Wei Wang. 2020.
\newblock Insrl: A multi-view learning framework fusing multiple information
  sources for distantly-supervised relation extraction.
\newblock \emph{arXiv preprint arXiv:2012.09370}.

\bibitem[{Chung et~al.(2022)Chung, Hou, Longpre, Zoph, Tay, Fedus, Li, Wang,
  Dehghani, Brahma et~al.}]{chung2022scaling}
Hyung~Won Chung, Le~Hou, Shayne Longpre, Barret Zoph, Yi~Tay, William Fedus,
  Eric Li, Xuezhi Wang, Mostafa Dehghani, Siddhartha Brahma, et~al. 2022.
\newblock Scaling instruction-finetuned language models.
\newblock \emph{arXiv preprint arXiv:2210.11416}.

\bibitem[{Devlin et~al.(2019)Devlin, Chang, Lee, and
  Toutanova}]{devlin2018bert}
Jacob Devlin, Ming-Wei Chang, Kenton Lee, and Kristina Toutanova. 2019.
\newblock \href {https://doi.org/10.18653/v1/N19-1423} {{BERT}: Pre-training of
  deep bidirectional transformers for language understanding}.
\newblock In \emph{Proceedings of the 2019 Conference of the North {A}merican
  Chapter of the Association for Computational Linguistics: Human Language
  Technologies, Volume 1 (Long and Short Papers)}, pages 4171--4186,
  Minneapolis, Minnesota. Association for Computational Linguistics.

\bibitem[{Dixit and Al-Onaizan(2019)}]{dixit2019span}
Kalpit Dixit and Yaser Al-Onaizan. 2019.
\newblock \href {https://doi.org/10.18653/v1/P19-1525} {Span-level model for
  relation extraction}.
\newblock In \emph{Proceedings of the 57th Annual Meeting of the Association
  for Computational Linguistics}, pages 5308--5314, Florence, Italy.
  Association for Computational Linguistics.

\bibitem[{Ellis et~al.(2012)Ellis, Li, Griffitt, Strassel, and
  Wright}]{ellis2012linguistic}
Joe Ellis, Xuansong Li, Kira Griffitt, Stephanie~M Strassel, and Jonathan
  Wright. 2012.
\newblock Linguistic resources for 2013 knowledge base population evaluations.
\newblock In \emph{TAC}.

\bibitem[{Gao et~al.(2023)Gao, Zhao, Yu, and Xu}]{gao2023exploring}
Jun Gao, Huan Zhao, Changlong Yu, and Ruifeng Xu. 2023.
\newblock Exploring the feasibility of chatgpt for event extraction.
\newblock \emph{arXiv preprint arXiv:2303.03836}.

\bibitem[{Hoffmann et~al.(2011)Hoffmann, Zhang, Ling, Zettlemoyer, and
  Weld}]{Hoffmann2011KnowledgeBasedWS}
R.~Hoffmann, Congle Zhang, Xiao Ling, Luke Zettlemoyer, and Daniel~S. Weld.
  2011.
\newblock Knowledge-based weak supervision for information extraction of
  overlapping relations.
\newblock In \emph{Proceedings of ACL}.

\bibitem[{Hu et~al.(2021)Hu, Shen, Wallis, Allen-Zhu, Li, Wang, Wang, and
  Chen}]{hu2021lora}
Edward~J Hu, Yelong Shen, Phillip Wallis, Zeyuan Allen-Zhu, Yuanzhi Li, Shean
  Wang, Lu~Wang, and Weizhu Chen. 2021.
\newblock Lora: Low-rank adaptation of large language models.
\newblock \emph{arXiv preprint arXiv:2106.09685}.

\bibitem[{Jeblick et~al.(2023)Jeblick, Schachtner, Dexl, Mittermeier,
  St{\"u}ber, Topalis, Weber, Wesp, Sabel, Ricke et~al.}]{jeblick2023chatgpt}
Katharina Jeblick, Balthasar Schachtner, Jakob Dexl, Andreas Mittermeier,
  Anna~Theresa St{\"u}ber, Johanna Topalis, Tobias Weber, Philipp Wesp,
  Bastian~Oliver Sabel, Jens Ricke, et~al. 2023.
\newblock Chatgpt makes medicine easy to swallow: an exploratory case study on
  simplified radiology reports.
\newblock \emph{European Radiology}, pages 1--9.

\bibitem[{Ji et~al.(2020)Ji, Yu, Li, Ma, Wu, Tan, and Liu}]{ji2020span}
Bin Ji, Jie Yu, Shasha Li, Jun Ma, Qingbo Wu, Yusong Tan, and Huijun Liu. 2020.
\newblock \href {https://doi.org/10.18653/v1/2020.coling-main.8} {Span-based
  joint entity and relation extraction with attention-based span-specific and
  contextual semantic representations}.
\newblock In \emph{Proceedings of the 28th International Conference on
  Computational Linguistics}, pages 88--99, Barcelona, Spain (Online).
  International Committee on Computational Linguistics.

\bibitem[{Jiang et~al.(2023)Jiang, Ren, and Lin}]{jiang-etal-2023-llm}
Dongfu Jiang, Xiang Ren, and Bill~Yuchen Lin. 2023.
\newblock \href {https://doi.org/10.18653/v1/2023.acl-long.792} {{LLM}-blender:
  Ensembling large language models with pairwise ranking and generative
  fusion}.
\newblock In \emph{Proceedings of the 61st Annual Meeting of the Association
  for Computational Linguistics (Volume 1: Long Papers)}, pages 14165--14178,
  Toronto, Canada. Association for Computational Linguistics.

\bibitem[{Jiang et~al.(2020)Jiang, Liu, Zhang, Yang, Xiao, and
  Wang}]{jiang2020surface}
Haiyun Jiang, JunTao Liu, Sheng Zhang, Deqing Yang, Yanghua Xiao, and Wei Wang.
  2020.
\newblock Surface pattern-enhanced relation extraction with global constraints.
\newblock \emph{Knowledge and Information Systems}, 62(12):4509--4540.

\bibitem[{Lang et~al.(2022)Lang, Agrawal, Kim, and Sontag}]{lang2022co}
Hunter Lang, Monica~N Agrawal, Yoon Kim, and David Sontag. 2022.
\newblock \href {https://proceedings.mlr.press/v162/lang22a.html} {Co-training
  improves prompt-based learning for large language models}.
\newblock In \emph{Proceedings of the 39th International Conference on Machine
  Learning}, volume 162 of \emph{Proceedings of Machine Learning Research},
  pages 11985--12003. PMLR.

\bibitem[{Leviathan et~al.(2023)Leviathan, Kalman, and
  Matias}]{leviathan2023fast}
Yaniv Leviathan, Matan Kalman, and Yossi Matias. 2023.
\newblock \href {https://proceedings.mlr.press/v202/leviathan23a.html} {Fast
  inference from transformers via speculative decoding}.
\newblock In \emph{Proceedings of the 40th International Conference on Machine
  Learning}, volume 202 of \emph{Proceedings of Machine Learning Research},
  pages 19274--19286. PMLR.

\bibitem[{Li et~al.(2019)Li, Yin, Sun, Li, Yuan, Chai, Zhou, and
  Li}]{li2019entity}
Xiaoya Li, Fan Yin, Zijun Sun, Xiayu Li, Arianna Yuan, Duo Chai, Mingxin Zhou,
  and Jiwei Li. 2019.
\newblock \href {https://doi.org/10.18653/v1/P19-1129} {Entity-relation
  extraction as multi-turn question answering}.
\newblock In \emph{Proceedings of the 57th Annual Meeting of the Association
  for Computational Linguistics}, pages 1340--1350, Florence, Italy.
  Association for Computational Linguistics.

\bibitem[{Ling and Weld(2012)}]{ling2012fine}
Xiao Ling and Daniel~S Weld. 2012.
\newblock Fine-grained entity recognition.
\newblock In \emph{Twenty-Sixth AAAI Conference on Artificial Intelligence}.

\bibitem[{Liu et~al.(2017)Liu, Ren, Zhu, Zhi, Gui, Ji, and
  Han}]{liu2017heterogeneous}
Liyuan Liu, Xiang Ren, Qi~Zhu, Shi Zhi, Huan Gui, Heng Ji, and Jiawei Han.
  2017.
\newblock Heterogeneous supervision for relation extraction: A representation
  learning approach.
\newblock \emph{arXiv preprint arXiv:1707.00166}.

\bibitem[{Liu et~al.(2019)Liu, Ott, Goyal, Du, Joshi, Chen, Levy, Lewis,
  Zettlemoyer, and Stoyanov}]{liu2019roberta}
Yinhan Liu, Myle Ott, Naman Goyal, Jingfei Du, Mandar Joshi, Danqi Chen, Omer
  Levy, Mike Lewis, Luke Zettlemoyer, and Veselin Stoyanov. 2019.
\newblock Roberta: A robustly optimized bert pretraining approach.
\newblock \emph{arXiv preprint arXiv:1907.11692}.

\bibitem[{OpenAI(2023)}]{openai2023gpt4}
OpenAI. 2023.
\newblock \href {http://arxiv.org/abs/2303.08774} {Gpt-4 technical report}.

\bibitem[{Ren et~al.(2017)Ren, Wu, He, Qu, Voss, Ji, Abdelzaher, and
  Han}]{ren2017cotype}
Xiang Ren, Zeqiu Wu, Wenqi He, Meng Qu, Clare~R Voss, Heng Ji, Tarek~F
  Abdelzaher, and Jiawei Han. 2017.
\newblock Cotype: Joint extraction of typed entities and relations with
  knowledge bases.
\newblock In \emph{Proceedings of the 26th International Conference on World
  Wide Web}, pages 1015--1024.

\bibitem[{Riedel et~al.(2010)Riedel, Yao, and McCallum}]{riedel2010modeling}
Sebastian Riedel, Limin Yao, and Andrew McCallum. 2010.
\newblock Modeling relations and their mentions without labeled text.
\newblock In \emph{Joint European Conference on Machine Learning and Knowledge
  Discovery in Databases}, pages 148--163. Springer.

\bibitem[{Sainz et~al.(2022)Sainz, Qiu, Lopez~de Lacalle, Agirre, and
  Min}]{sainz-etal-2022-zs4ie}
Oscar Sainz, Haoling Qiu, Oier Lopez~de Lacalle, Eneko Agirre, and Bonan Min.
  2022.
\newblock \href {https://doi.org/10.18653/v1/2022.naacl-demo.4} {{ZS}4{IE}: A
  toolkit for zero-shot information extraction with simple verbalizations}.
\newblock In \emph{Proceedings of the 2022 Conference of the North American
  Chapter of the Association for Computational Linguistics: Human Language
  Technologies: System Demonstrations}, pages 27--38, Hybrid: Seattle,
  Washington + Online. Association for Computational Linguistics.

\bibitem[{Su et~al.(2021)Su, Lu, Pan, Wen, and Liu}]{su2021roformer}
Jianlin Su, Yu~Lu, Shengfeng Pan, Bo~Wen, and Yunfeng Liu. 2021.
\newblock Roformer: Enhanced transformer with rotary position embedding.
\newblock \emph{arXiv preprint arXiv:2104.09864}.

\bibitem[{Takanobu et~al.(2019)Takanobu, Zhang, Liu, and
  Huang}]{takanobu2019hierarchical}
Ryuichi Takanobu, Tianyang Zhang, Jiexi Liu, and Minlie Huang. 2019.
\newblock A hierarchical framework for relation extraction with reinforcement
  learning.
\newblock In \emph{Proceedings of AAAI}, volume~33, pages 7072--7079.

\bibitem[{Tang et~al.(2023)Tang, Han, Jiang, and Hu}]{tang2023does}
Ruixiang Tang, Xiaotian Han, Xiaoqian Jiang, and Xia Hu. 2023.
\newblock Does synthetic data generation of llms help clinical text mining?
\newblock \emph{arXiv preprint arXiv:2303.04360}.

\bibitem[{Touvron et~al.(2023{\natexlab{a}})Touvron, Lavril, Izacard, Martinet,
  Lachaux, Lacroix, Rozi{\`e}re, Goyal, Hambro, Azhar
  et~al.}]{touvron2023llama}
Hugo Touvron, Thibaut Lavril, Gautier Izacard, Xavier Martinet, Marie-Anne
  Lachaux, Timoth{\'e}e Lacroix, Baptiste Rozi{\`e}re, Naman Goyal, Eric
  Hambro, Faisal Azhar, et~al. 2023{\natexlab{a}}.
\newblock Llama: Open and efficient foundation language models.
\newblock \emph{arXiv preprint arXiv:2302.13971}.

\bibitem[{Touvron et~al.(2023{\natexlab{b}})Touvron, Martin, Stone, Albert,
  Almahairi, Babaei, Bashlykov, Batra, Bhargava, Bhosale
  et~al.}]{touvron2023llama2}
Hugo Touvron, Louis Martin, Kevin Stone, Peter Albert, Amjad Almahairi, Yasmine
  Babaei, Nikolay Bashlykov, Soumya Batra, Prajjwal Bhargava, Shruti Bhosale,
  et~al. 2023{\natexlab{b}}.
\newblock Llama 2: Open foundation and fine-tuned chat models.
\newblock \emph{arXiv preprint arXiv:2307.09288}.

\bibitem[{Wadhwa et~al.(2023)Wadhwa, Amir, and
  Wallace}]{wadhwa-etal-2023-revisiting}
Somin Wadhwa, Silvio Amir, and Byron Wallace. 2023.
\newblock \href {https://doi.org/10.18653/v1/2023.acl-long.868} {Revisiting
  relation extraction in the era of large language models}.
\newblock In \emph{Proceedings of the 61st Annual Meeting of the Association
  for Computational Linguistics (Volume 1: Long Papers)}, pages 15566--15589,
  Toronto, Canada. Association for Computational Linguistics.

\bibitem[{Wang et~al.(2020)Wang, Yu, Zhang, Liu, Zhu, and
  Sun}]{wang2020tplinker}
Yucheng Wang, Bowen Yu, Yueyang Zhang, Tingwen Liu, Hongsong Zhu, and Limin
  Sun. 2020.
\newblock \href {https://doi.org/10.18653/v1/2020.coling-main.138} {{TPL}inker:
  Single-stage joint extraction of entities and relations through token pair
  linking}.
\newblock In \emph{Proceedings of the 28th International Conference on
  Computational Linguistics}, pages 1572--1582, Barcelona, Spain (Online).
  International Committee on Computational Linguistics.

\bibitem[{Wei et~al.(2021)Wei, Bosma, Zhao, Guu, Yu, Lester, Du, Dai, and
  Le}]{wei2021finetuned}
Jason Wei, Maarten Bosma, Vincent~Y Zhao, Kelvin Guu, Adams~Wei Yu, Brian
  Lester, Nan Du, Andrew~M Dai, and Quoc~V Le. 2021.
\newblock Finetuned language models are zero-shot learners.
\newblock \emph{arXiv preprint arXiv:2109.01652}.

\bibitem[{Wei et~al.(2023)Wei, Cui, Cheng, Wang, Zhang, Huang, Xie, Xu, Chen,
  Zhang et~al.}]{wei2023zero}
Xiang Wei, Xingyu Cui, Ning Cheng, Xiaobin Wang, Xin Zhang, Shen Huang, Pengjun
  Xie, Jinan Xu, Yufeng Chen, Meishan Zhang, et~al. 2023.
\newblock Zero-shot information extraction via chatting with chatgpt.
\newblock \emph{arXiv preprint arXiv:2302.10205}.

\bibitem[{Wei et~al.(2020)Wei, Su, Wang, Tian, and Chang}]{wei2020novel}
Zhepei Wei, Jianlin Su, Yue Wang, Yuan Tian, and Yi~Chang. 2020.
\newblock \href {https://doi.org/10.18653/v1/2020.acl-main.136} {A novel
  cascade binary tagging framework for relational triple extraction}.
\newblock In \emph{Proceedings of the 58th Annual Meeting of the Association
  for Computational Linguistics}, pages 1476--1488, Online. Association for
  Computational Linguistics.

\bibitem[{Xie et~al.(2021)Xie, Liang, Liu, Huang, Huang, and
  Xiao}]{xie2021revisiting}
Chenhao Xie, Jiaqing Liang, Jingping Liu, Chengsong Huang, Wenhao Huang, and
  Yanghua Xiao. 2021.
\newblock \href {https://doi.org/10.18653/v1/2021.acl-long.277} {Revisiting the
  negative data of distantly supervised relation extraction}.
\newblock In \emph{Proceedings of the 59th Annual Meeting of the Association
  for Computational Linguistics and the 11th International Joint Conference on
  Natural Language Processing (Volume 1: Long Papers)}, pages 3572--3581,
  Online. Association for Computational Linguistics.

\bibitem[{Xu et~al.(2023)Xu, Xu, Wang, Liu, Zhu, and McAuley}]{xu2023small}
Canwen Xu, Yichong Xu, Shuohang Wang, Yang Liu, Chenguang Zhu, and Julian
  McAuley. 2023.
\newblock Small models are valuable plug-ins for large language models.
\newblock \emph{arXiv preprint arXiv:2305.08848}.

\bibitem[{Yu et~al.(2020)Yu, Zhang, Shu, Wang, Liu, Wang, and
  Li}]{yu2020jointer}
Bowen Yu, Zhenyu Zhang, Xiaobo Shu, Yubin Wang, Tingwen Liu, Bin Wang, and
  Sujian Li. 2020.
\newblock Joint extraction of entities and relations based on a novel
  decomposition strategy.
\newblock In \emph{Proceedings of ECAI}.

\bibitem[{Yuan et~al.(2023)Yuan, Xie, and Ananiadou}]{yuan-etal-2023-zero}
Chenhan Yuan, Qianqian Xie, and Sophia Ananiadou. 2023.
\newblock \href {https://doi.org/10.18653/v1/2023.bionlp-1.7} {Zero-shot
  temporal relation extraction with {C}hat{GPT}}.
\newblock In \emph{The 22nd Workshop on Biomedical Natural Language Processing
  and BioNLP Shared Tasks}, pages 92--102, Toronto, Canada. Association for
  Computational Linguistics.

\bibitem[{Zheng et~al.(2023)Zheng, Chiang, Sheng, Zhuang, Wu, Zhuang, Lin, Li,
  Li, Xing et~al.}]{zheng2023judging}
Lianmin Zheng, Wei-Lin Chiang, Ying Sheng, Siyuan Zhuang, Zhanghao Wu, Yonghao
  Zhuang, Zi~Lin, Zhuohan Li, Dacheng Li, Eric Xing, et~al. 2023.
\newblock Judging llm-as-a-judge with mt-bench and chatbot arena.
\newblock \emph{arXiv preprint arXiv:2306.05685}.

\end{thebibliography}
% \bibliography{custom}

% \section{Language Resource References}
% \label{lr:ref}
% \bibliographystylelanguageresource{lrec-coling2024-natbib}
% \bibliographylanguageresource{languageresource}

\newpage
\appendix

\section{Dataset Introduction}
\label{sec:dataset}
\paragraph*{NYT series}
NYT is based on the articles in New York Times. There are many derived datasets with better labeling. NYT10~\cite{riedel2010modeling} and NYT11~\cite{Hoffmann2011KnowledgeBasedWS} label the complete entities. Moreover, NYT10-HRL and NYT11-HRL~\cite{takanobu2019hierarchical} are better versions that are processed by optimizing the relation labels.

\paragraph*{HacRED} HacRED~\cite{hacred}\footnote{\url{https://github.com/qiaojiim/hacred}} is a novel challenging extraction dataset. It analyzes the performance gap between popular datasets and practical applications, and carefully selects and designs more hard cases. HacRED consists of 65,225 relational facts annotated from 9,231 wiki documents with sufficient and diverse hard cases, which poses a very high challenge to many current complex extraction methods.

\paragraph*{SKE21}
SKE19\footnote{\url{http://ai.baidu.com/broad/download?dataset=sked}} is published by Baidu, and is currently the largest dataset available for complex relational triple extraction. Since its testing set is unpublished, and there are some errors in the validation set, a version named SKE21 is published by \citet{xie2021revisiting}. The testing set of SKE21 is carefully manually relabeled and contains 1,150 sentences and 2,765 annotated triples.

\paragraph*{Wiki-KBP}
Wiki-KBP \cite{ling2012fine} is based on the articles in Wikipedia. There're 1.5M sentences in training set which are automatically labeled using distant supervision and handcrafted patterns by \cite{liu2017heterogeneous}, and the test set contains 289 sentences selected by the author of \cite{ren2017cotype} from the manual annotations in 2013 KBP slot filling results \cite{ellis2012linguistic}.

% 数据集介绍，实验细节（超参数等），prompt模板
\section{Experiment Details}
\label{sec:experiments}
Our experiments are conducted on two A800 GPUs. All deep models, including the LLMs and the evaluation model, are fine-tuned or implemented using the PyTorch framework. We employed AdamW optimizer as the optimizer. For the evaluation model, we first initialize the model with bert-base-cased and chinese-roberta-wwm-ext respectively, then train 20 epochs in English corpus task, and 40 epochs in Chinese. For the fine-tuning of LLMs, we randomly select 1500 items from the train set for each dataset and train 30 epochs. Our codes and hyper-parameters can be found at \url{https://github.com/Ding-Papa/Evaluating-filtering-coling24}.

\section{Instruction Template}
\label{sec:template}
Here we provide the instruction templates that guide the LLMs for relational triple extraction. First is the template for directly using the LLMs to perform extraction, i.e., the first stage of our method.

\noindent\textbf{Template for the first stage}: \\
\noindent\texttt{Pre-define the following relation list $r$, please extract all triples containing the above relations from the given sentence $S$.} \\
\noindent\texttt{Note that the relation name of the triple must be selected from the above list, and other relations not listed are not considered. Please output according to the specified format: [\{"s": subject1, "o": object1, "p": relation1\}, \{"s": subject2, "o": object2, "p": relation2\},...]} \\
\noindent\texttt{(Optional) Here are some examples: ...} \\
\noindent\texttt{Now given the following input, please complete the extracting task.} \\
\noindent\texttt{Please output as many triples as possible that meet the requirements.} \\
\noindent\texttt{Input: } $S_i,r_i$
\\

In the second stage, the input of the LLMs consists of the first-stage extraction results and candidate pairs extracted by the evaluation model. The LLMs are prompted to recheck the original results, assign relations to the appropriate candidate entity pairs, and output the final extracted triples to complete the extraction.

\noindent\textbf{Template for the second stage}: \\
\noindent\texttt{Pre-define the following relation list $r$. We want to extract all triples containing the above relations from the given sentence $S$. Here are the original extraction results $A$.} \\
\noindent\texttt{Now we claim that the entity pairs that may be related in the above sentence are $(s_1,o_1),(s_2,o_2),...$} \\
\noindent\texttt{Please check the original results and fill in the missing triples, remove the wrong triples and output the final results.} \\
\noindent\texttt{Constraints and output format are the same as stage 1.} \\
\noindent\texttt{Please output according to the specified format.} \\

\end{document}